%% file: main.tex
\documentclass{article}
\usepackage{iclr2027/iclr2027_conference,times}
\usepackage[T1]{fontenc}

\usepackage{hyperref}
\usepackage{url}
\usepackage{booktabs}
\usepackage{amsmath}
\usepackage{amssymb}
\usepackage{graphicx}
\usepackage{xcolor}
\usepackage{wrapfig}
\usepackage{placeins}
\usepackage{needspace}

\input{numbers}

\input{result_status}
\input{numbers_timing}

\newcommand{\pending}[1]{\textcolor{gray}{[pending: #1]}}

\title{\raggedright Do Temporal Link Predictors Need Learned Memory?
A Smoothed-Count Baseline with a Handful of Parameters}

\iclrfinalcopy

\author{Lisi Qarkaxhija \quad Ingo Scholtes\\
Chair of Machine Learning for Complex Networks\\
Center for Artificial Intelligence and Data Science (CAIDAS)\\
Julius-Maximilians-Universit\"at W\"urzburg, DE\\
\texttt{name.surname@uni-wuerzburg.de}}

\begin{document}
\maketitle

\begin{abstract}
Many temporal link predictors summarize past interactions through learned node representations.
We examine whether simple counts of recurring interaction patterns can provide competitive predictions without learning these representations.
We propose a temporal link predictor based on statistical language modelling.
It pools transition and co-occurrence counts across sources to predict links that a source has never formed.
We smooth sparse estimates using destination frequencies or Kneser-Ney continuation counts.
A shared log-linear rule combines these estimates with popularity, source history, and recency, without node embeddings.
In our main evaluation, the model achieves the highest MRR among the compared methods on \NBestAll{} out of \NDatasets{} datasets from TGB and TGB-Seq.
It also outperforms EdgeBank and Base3 on all \NDatasets{} datasets and the heuristic family on \NLearnedAhead{}.
These gains extend to datasets designed to limit repeated edges.
With only \NParams{}--\NParamsFull{} learned parameters, our model provides a simple and competitive baseline for evaluating future neural temporal link predictors.

\end{abstract}

\section{Introduction}
\label{sec:intro}

Temporal link prediction models learn from past interactions to predict future links.
Many methods represent this history through learned node states, embeddings, or neighborhood encodings \citep{kumar2019jodie, trivedi2019dyrep, rossi2020tgn, wang2021cawn, yu2023dygformer, lu2024tpnet}.
Simple temporal patterns also carry predictive information.
Edges repeat, and destination frequencies change over time \citep{poursafaei2022edgebank, daniluk2023global}.
Recent work therefore questions whether complex architectures are needed for temporal prediction \citep{cong2023graphmixer}.
Strong statistical baselines help measure the predictive gains from learned representations.

Existing count-based methods use only part of this information.
EdgeBank predicts the recurrence of previously observed edges \citep{poursafaei2022edgebank}.
PopTrack ranks destinations by decayed frequency \citep{daniluk2023global}.
Other methods combine recency, popularity, and co-occurrence through fixed rules \citep{cornell2025heuristics, kondrup2025base3}.
These methods have mainly been studied on datasets with repeated edges.
Their predictive value when sources rarely revisit a destination remains less understood.
TGB-Seq tests this setting \citep{yi2025tgbseq}.

Temporal interactions provide recurring prediction contexts.
The recent destinations of a source can inform its next destination, but individual histories often contain few observations of each transition.
Statistical language modelling addresses this problem by pooling observations of recurring contexts and smoothing sparse estimates toward a broader distribution \citep{katz1987estimation, chen1999empirical}.

We use this construction to build a temporal link predictor from smoothed counts and recency features, with a shared log-linear scoring rule.
We consider two backoff distributions for contexts with few observations.
Marginal backoff uses destination frequency, favoring items that are common overall.
This follows the use of unigram frequencies in language-model interpolation \citep{mackay1995hierarchical}.
Kneser-Ney continuation backoff uses the number of distinct preceding destinations, favoring items observed across many transition contexts \citep{kneser1995improved}.

\begin{figure}[t]
\centering
\includegraphics[width=\textwidth]{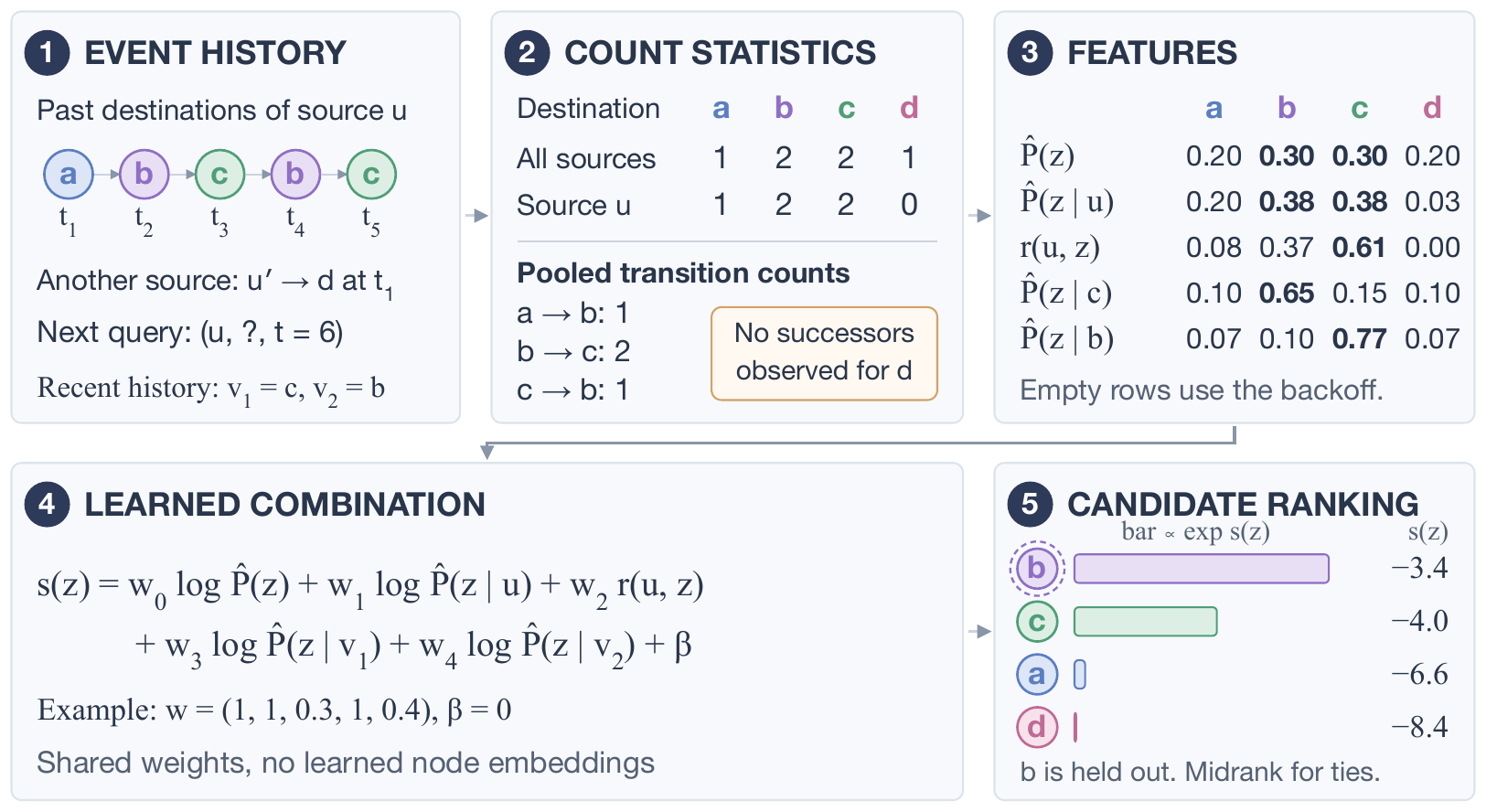}
\caption{From six observed interactions to candidate ranking.
The reference configuration uses $K=2$, $\alpha=1$, undecayed counts, and marginal backoff.
We set $t_i=i$, query at $t=6$, and use recency scale $\tau_r T=2$.
Panels 1--3 form the features.
Panels 4--5 apply illustrative weights and rank candidates.
Appendix~\ref{app:optionals} gives the calculations and further features.}
\label{fig:hero}
\end{figure}

A destination can receive support from several statistics (Figure~\ref{fig:hero}).
It may be popular overall, frequent in the history of the source, or recently visited.
It may also commonly follow a recent destination of the source in the histories of other sources.
Transitions measure this relation between consecutive interactions.
Windowed co-occurrence also includes destinations separated by intervening events.
Both can support a link that the source has never formed.
The model stores decayed counts and recent destination lists.
It learns only the weights used to combine the resulting estimates.
The model has \NParams{}--\NParamsFull{} learned parameters at context depth $K=\KLags$, shared across all nodes, and no learned node embeddings.
Fixed update rules allow us to compute training features once and reuse them across epochs.

Across \NDatasets{} TGB and TGB-Seq datasets, the model exceeds EdgeBank and Base3 on all datasets and the heuristic family on \NLearnedAhead{} under matched event-wise\iftimingready{} and batch-wise\fi{} updates.
The coverage analysis shows that source-specific transition contexts can be absent while pooled transition evidence remains available.
This distinction helps explain prediction beyond repeated source-specific patterns.

\textbf{Contributions.}
\begin{itemize}
\item \textbf{A statistical model of interaction history.}
A log-linear model combines smoothed temporal counts and recency features, with weights shared across nodes and no learned node embeddings (Section~\ref{sec:model}).
\item \textbf{Evaluation across two benchmark protocols.}
On \NDatasets{} TGB and TGB-Seq datasets, we compare against recomputed count baselines and reported graph and sequential models (Section~\ref{sec:experiments}).
\item \textbf{Analysis of statistical context.}
Controlled comparisons measure the effects of context depth, co-occurrence, and backoff distributions.
Context coverage distinguishes source-specific transition evidence from evidence pooled across sources, including where the former is absent (Sections~\ref{sec:backoff}--\ref{sec:coverage}).
\end{itemize}

\section{Related work}
\label{sec:related}

\paragraph{Graph-based models.}
JODIE, DyRep, and TGN learn recurrent node states that evolve with interactions \citep{kumar2019jodie, trivedi2019dyrep, rossi2020tgn}.
CAWN and DyGFormer encode temporal walks and neighborhood sequences \citep{wang2021cawn, yu2023dygformer}.
DBGNNs apply message passing to De Bruijn graphs that capture higher-order dependencies in time-respecting walks \citep{qarkaxhija2022dbgnn}.
TPNet combines recent interactions with time-decayed temporal walk counts \citep{lu2024tpnet}.
GraphMixer shows that simpler encoders can also perform well \citep{cong2023graphmixer}.
These temporal models learn how to encode interaction history.
For static graphs, \citet{qarkaxhija2025untrained} show that removing trainable parameters from message passing layers can preserve or improve link prediction performance.
Our model represents interaction history with smoothed counts and learns a shared scoring rule.

\paragraph{Count-based methods.}
EdgeBank predicts whether an edge has occurred before \citep{poursafaei2022edgebank}.
PopTrack uses decayed destination frequencies \citep{daniluk2023global}.
\citet{cornell2025heuristics} combine local and global recency and popularity through a fixed priority order.
Base3 interpolates EdgeBank, PopTrack, and a co-occurrence memory \citep{kondrup2025base3}.
Our model adds smoothed conditional estimates and learns their contribution to the score.

\paragraph{Sequential recommenders.}
Session-based methods predict the next item from recent interactions.
SKNN uses session co-occurrence \citep{ludewig2018sknn}.
SGNN-HN propagates information through a session graph with learned item embeddings \citep{pan2020sgnnhn}.
CRAFT uses cross-attention between each candidate and the recent interactions of the source \citep{yi2026future}.
SGNN-HN and CRAFT provide the strongest reported TGB-Seq scores used in our comparison.
Their use of recent sequences motivates our comparison with pooled transition and co-occurrence counts.

\paragraph{Evaluation.}
Temporal link prediction scores depend on the evaluation protocol.
Negative sampling can change method rankings \citep{poursafaei2022edgebank, cornell2025measuring}, sampled metrics can reverse the ordering obtained from full rankings \citep{krichene2020sampled}, and batch evaluation can withhold recent interactions \citep{lampert2026forecasting}.
Analyses of learned models and simple baselines motivate closer control of these factors \citep{hayes2025learn, bechlerspeicher2025position}.
We recompute count baselines on the benchmark candidates and measure sensitivity to history-update timing.
Context coverage complements the edge-based surprise index of \citet{poursafaei2022edgebank} by measuring whether a transition context has prior observations.

\paragraph{Statistical language modelling.}
Interpolation and backoff combine estimates from contexts with different amounts of evidence \citep{katz1987estimation, chen1999empirical}.
Unigram interpolation supplies background probabilities from overall word frequencies \citep{mackay1995hierarchical}.
Kneser-Ney smoothing uses the number of distinct preceding contexts to estimate a continuation distribution \citep{kneser1995improved}.
\citet{chen1999empirical} found this approach effective in language modelling.
In temporal link prediction, the corresponding counts describe source histories and destination transitions, with exponential decay accounting for their age.

\section{Problem setting and evaluation}
\label{sec:setting}

An interaction stream consists of events $(u, v, t)$, where $u$ is a source, $v$ a destination, and $t$ a timestamp.
At a query $(u, t)$, the task is to rank the observed destination above the benchmark negative candidates.
Both benchmarks report mean reciprocal rank (MRR) over test queries.
We follow their midrank tie rule.
The observed destination has rank $1 + n_{>} + \tfrac{1}{2}n_{=}$, where $n_{>}$ counts negatives with a higher score and $n_{=}$ counts negatives with the same score.
Only the rank of the observed destination enters MRR.
This treatment matters for count models, which often assign equal scores to unobserved candidates.

TGB-Seq \citep{yi2025tgbseq} provides 100 uniformly sampled negative destinations for each test event.
TGB \citep{huang2023tgb} uses dataset-specific candidate sets.
For example, \texttt{tgbl-wiki} uses all possible negative destinations, whereas \texttt{tgbl-review} uses 100 sampled negatives per positive edge.
Historical-random sampling includes destinations visited by the source during training.
Historical negatives test whether a model can distinguish the positive from previously visited destinations.
Because candidate distributions differ, we analyse each benchmark separately.

We use the splits, negative candidates, and MRR evaluation supplied by TGB and TGB-Seq.
We compare with reported neural benchmark results and recompute the count baselines.
Appendix~\ref{app:datasets} documents the score sources and evaluation details.

\paragraph{Notation.}
Throughout, $u$ denotes the source of the current query and $z$ a candidate destination from the destination vocabulary $\mathcal{V}$.
The $K$ most recent destinations of $u$ at query time are $v_1, \dots, v_K$, ordered from most recent.
The vocabulary is fixed for each dataset and includes candidate destinations first observed after training.
$t$ is the query time, and $T$ is the duration of the training split.
Probabilities written with a hat, such as $\hat{P}(z \mid u)$, are smoothed estimates (Section~\ref{sec:model}).

\section{A statistical model for temporal link prediction}
\label{sec:model}

The model represents history through count tables and recent destination lists.
Count estimates describe how often a destination occurs in a particular context.
Recency features describe how recently it was visited.
A shared log-linear rule combines these two kinds of evidence.
The stored counts, timestamps, and destination lists follow fixed update rules, and learning determines the contribution of each feature to the score.

\subsection{From complete histories to recurring contexts}
\label{sec:contexts}

Complete interaction histories rarely repeat, leaving little evidence for estimating their conditional destination distributions.
We use shorter contexts that recur across queries.
The marginal $\hat{P}(z)$ estimates destination frequency.
The source conditional $\hat{P}(z\mid u)$ estimates the choices of a particular source.
We use the last $K$ destinations visited by the source.
For each destination $v_k$, the transition conditional $\hat{P}(z\mid v_k)$ estimates what immediately follows it using transitions observed across all sources.
Windowed co-occurrence broadens this context to include pairs separated by intervening interactions.
It pools counts over the recent destinations to form the estimate $\hat{P}_{\mathrm{co}}(z\mid v_1,\ldots,v_K)$.

In film recommendation, suppose Alice recently watched films $A$ and $B$.
The marginal estimate describes how popular a film is across all viewers.
The source conditional describes how often Alice has watched it.
The transition conditionals use what all viewers watched immediately after $A$ or $B$ to estimate her next choice.
Co-occurrence also counts films watched within a few choices after $A$ or $B$.
These estimates use different parts of the same history.

\subsection{Streaming count state}

The count state records destination frequencies, source histories, transitions, and windowed co-occurrences.
Each source also retains its recent destination list.
Destination counts $c(z; t)$ measure decayed popularity, as in PopTrack \citep{daniluk2023global}.
Their total is $C(t)=\sum_{z\in\mathcal{V}}c(z;t)$.
Source-destination counts $c(u,z; t)$ record pair history together with the last interaction time.
Their keys form the edge memory used by EdgeBank \citep{poursafaei2022edgebank}.
Transition counts $c(v,z; t)$ record how often a source moves from $v$ to $z$ in consecutive interactions, pooled across all sources.
At query time, each of the last $K$ destinations selects a separate row from this transition table.
The model combines these estimates without treating the full sequence as one context.

Windowed co-occurrence counts $c_{\mathrm{co}}(v,z;t)$ record how often $z$ follows $v$ within the last $K$ interactions of the same source.
These counts are also pooled across sources.
An event updates the count for each destination in the recent window, including repeated occurrences.
The table can therefore contain pairs that never occur consecutively, as in session-based recommendation \citep{ludewig2018sknn}.

Tables store only keys that have occurred.
Missing keys have count zero.
Each event at time $t_i$ contributes $e^{-(t-t_i)/\tau}$ at query time $t$, and each count sums these contributions over its matching events.
Decay is applied lazily when a cell is accessed.
Updating popularity, source history, and consecutive transitions takes $O(1)$ operations per event.
Co-occurrence requires up to $K$ count updates.
Exponential decay preserves the relation between row entries and their totals, so the normalised estimators remain proper probabilities.
The time constant $\tau$ is set from timestamps and is not tuned (Appendix~\ref{app:hyper}).

\subsection{Smoothed conditional estimators}
\label{sec:smoothing}

For a context $x\in\{u,v_1,\dots,v_K\}$, let $n(x;t)=\sum_{z\in\mathcal{V}}c(x,z;t)$ denote its total decayed count.
Each context contributes one smoothed estimator.
We add pseudocounts from a base distribution $B$, using the Dirichlet smoothing form of \citet{mackay1995hierarchical}:
\begin{align}
\hat{P}(z \mid x) &= \frac{c(x, z; t) + \alpha\,B(z)}{n(x; t) + \alpha},
\label{eq:estimators}
\\
B(z) &\in \Big\{\;
\hat{P}(z) = \frac{c(z; t) + \alpha}{C(t) + \alpha\,\lvert\mathcal{V}\rvert},
\qquad
\hat{P}_{\mathrm{c}}(z) = \frac{N_{1+}(\cdot\, z) + \alpha}{\sum_{z'} N_{1+}(\cdot\, z') + \alpha\,\lvert\mathcal{V}\rvert}
\;\Big\},
\label{eq:bases}
\end{align}
with $\alpha = 1$, $\lvert\mathcal{V}\rvert$ the size of the destination vocabulary, and $N_{1+}(\cdot\, z) = \lvert\{v : c(v, z; t) > 0\}\rvert$ the number of distinct destinations $v$ after which $z$ has been observed.
A context with no observations satisfies $\hat{P}(z \mid x) = B(z)$ exactly, so an empty context uses the base distribution.

Equation~\ref{eq:bases} defines two backoff distributions.
The marginal $\hat{P}(z)$ measures how frequently $z$ occurs.
The continuation base $\hat{P}_{\mathrm{c}}(z)$ measures how many distinct destinations have preceded $z$.
A destination that occurs often after only one predecessor can therefore have high frequency but a continuation count of one.
The continuation count is the number of observed cells in column $z$ of the transition table.
These counts do not decay, so the base retains predecessor types from the full observed history.

The continuation counts come from Kneser-Ney smoothing \citep{kneser1995improved, chen1999empirical}.
The resulting base assigns more backoff probability to items observed in many contexts.
This matters most when the conditioning row has few observations.
Both bases use the same Dirichlet smoothing rule.
This isolates the effect of the backoff distribution.
The marginal also remains a separate feature under either base.

The co-occurrence estimate combines evidence from the whole recent window.
We sum the candidate counts over the corresponding rows, normalize by their total count, and smooth toward $B$.
For $L\leq K$ available history positions,
\begin{equation}
\hat{P}_{\mathrm{co}}(z\mid v_1,\ldots,v_L)
=\frac{\sum_{k=1}^{L}c_{\mathrm{co}}(v_k,z;t)+\alpha B(z)}
{\sum_{k=1}^{L}\sum_{z'}c_{\mathrm{co}}(v_k,z';t)+\alpha}.
\label{eq:cooccurrence}
\end{equation}
Repeated destinations contribute once per history position.
With no co-occurrence observations, Equation~\ref{eq:cooccurrence} returns $B(z)$.
Appendix~\ref{app:optionals} illustrates each estimator on a worked stream.

\subsection{Recency features}
\label{sec:recency}

Recency describes the time of the latest interaction with a candidate.
Source-specific recency uses the last visit by $u$: $r(u,z)=e^{-(t-t_{\mathrm{last}}(u,z))/(\tau_r T)}$, where $T$ is the training duration and $\tau_r$ is a fixed fraction.
It is zero if $u$ has never visited $z$.
In the film example, this feature measures how recently Alice watched the candidate film.

\emph{Global recency} measures how recently any source interacted with a candidate.
Its value is $e^{-(t-t_{\mathrm{last}}(z))/(\tau_r T)}$, where $t_{\mathrm{last}}(z)$ is the last interaction time of $z$ across all sources.
This is the global-recency signal used by \citet{cornell2025heuristics}.
It can favor a film watched recently by someone other than the current viewer.

\emph{Multi-scale recency} measures the same source-destination interaction as $r(u,z)$ at three fixed scales, $\tau_r/10$, $\tau_r$, and $10\tau_r$.
The middle scale is $r(u,z)$, so this representation contributes two further features.
Multiple scales allow the model to combine fast and slow recency effects.
This choice is motivated by retention curves that decay more slowly than a single exponential \citep{anderson1991reflections}.
The learned weights determine the influence of each scale.

\subsection{Learned combination}
\label{sec:score}

Let $\boldsymbol\phi(u,z,t)$ contain the log-probabilities from the count estimates and the untransformed recency values.
We score each candidate with a shared log-linear rule:
\begin{equation}
s(z)=\boldsymbol w^{\mathsf T}\boldsymbol\phi(u,z,t)+\beta.
\label{eq:score}
\end{equation}
Each feature has one learned weight, and $\beta$ is a learned bias.
The weights are unconstrained.
A negative weight makes a larger feature value evidence against a candidate.
Probability mixtures instead require nonnegative weights.

If a source has fewer than $K$ past interactions, transition features for unavailable history positions are zero for every candidate.
These inactive features do not affect the ranking.
An available destination with no observed successors instead uses the backoff distribution in Equation~\ref{eq:estimators}.
Figure~\ref{fig:hero} shows how individual features contribute to a candidate score.

The evaluated configurations have \NParams{}--\NParamsFull{} learned parameters at $K=\KLags$, including the bias.
Changing the backoff distribution adds no parameters.
There are no node or item embeddings, message passing, or learned per-node state.
A source enters through its recent destinations and count rows.
The same scoring rule applies to nodes first observed at test time.
The statistical state grows with the number of observed destinations and pairs, so the parameter count alone does not describe the memory cost.

\subsection{Fitting}
\label{sec:fitting}

Features are extracted in a single chronological pass, before adding each event to history.
The resulting vectors are reused across epochs and shuffled into minibatches to fit only the scoring weights.
Training minimises a sampled softmax loss with one positive and one uniformly sampled negative destination per query.
The number of negatives per positive follows TGN \citep[Appendix~A.4]{rossi2020tgn}.
Test candidates are provided by the benchmark.

\section{Results and discussion}
\label{sec:experiments}

\textbf{Setup.}
We evaluate on \NTGBDatasets{} TGB and \NSeqDatasets{} TGB-Seq datasets.
Main comparisons use five runs.
Feature, depth, and smoothing analyses use three runs.
Timing comparisons reuse the five main fits.
Tables report means and sample standard deviations.
Hyperparameters are fixed as specified in Section~\ref{sec:model} and Appendix~\ref{app:hyper}.
Tables~\ref{tab:tgb} and~\ref{tab:tgbseq} report TGB and TGB-Seq separately.

We fix one feature configuration per dataset and backoff distribution, as listed in Appendix~\ref{app:hyper}.
These feature sets were chosen on validation data before the reported fits.
Mean validation MRR over five runs chooses between the two backoff configurations.
We also choose the heuristic rule on validation MRR.
All performance rankings use test MRR.
For controlled comparisons, the \emph{reference configuration} contains popularity, source history, the $K$ transition estimates, and source-specific recency at scale $\tau_r$.
We add co-occurrence, global recency, or two further source-recency scales separately and together, under each backoff.
These comparisons use the same three seeds for both configurations.

Following the streaming setting of TGB, observed test interactions enter history for later predictions while learned weights remain fixed \citep[Appendix~C]{huang2023tgb}.
Our history updates use fixed count operations, making event-wise evaluation practical.
We update statistics after each scored event, retaining file order for tied timestamps.
Our model, EdgeBank, Base3, and the heuristic family use the same candidates and event-wise updates.
Appendix~\ref{app:timing} examines batch-wise and strict-timestamp updates.

\begin{table}[!t]
\centering
\caption{Test MRR (\%) on TGB.
Our entries are means and sample standard deviations over five runs.
EdgeBank is the stronger of its two variants.
Best neural gives the strongest reported neural benchmark score (Appendix~\ref{app:datasets}).
Bold marks the highest MRR per row.}
\label{tab:tgb}
\small
\input{tab_tgb}
\end{table}

\begin{table}[!t]
\centering
\caption{Test MRR (\%) on TGB-Seq.
Columns follow Table~\ref{tab:tgb}.
Count baselines use the TGB-Seq candidate sets.}
\label{tab:tgbseq}
\small
\input{tab_tgbseq}
\end{table}

The model achieves the highest MRR among all compared methods on \NBestAll{} of \NDatasets{} datasets.
Each of these leads holds in all five runs.
CRAFT and CRAFT-R together lead on \NBestCraftFamily{} datasets, the next highest count.
On TGB, the lead over TPNet on \texttt{tgbl-wiki} is \MarginWiki{} points.
The largest gains over reported TGB references occur on \texttt{tgbl-uci}, \texttt{tgbl-enron}, and \texttt{tgbl-lastfm}.
Recomputed heuristics are also strong on these datasets, and exceed our model on \texttt{tgbl-uci}.
On TGB-Seq, the model leads on \texttt{ML-20M}, \texttt{Flickr}, and \texttt{YouTube}.
These datasets have no previously observed source-specific transition contexts at test time (Section~\ref{sec:coverage}).

The model remains below the reported references on \NBehindTGB{} TGB and \NBehindSeq{} TGB-Seq datasets.
The largest deficits within each benchmark occur on \texttt{tgbl-comment} and \texttt{GoogleLocal}.
Against recomputed count methods, it exceeds EdgeBank and Base3 on \NBeatBaseThree{} of \NBaseThreeTotal{} datasets and the heuristic rule on \NLearnedAhead{} of \NLearnedTotal{}.\iftimingready{} These gains persist when all counting methods use batch-200 updates.
Under this schedule, our model still achieves the highest MRR in the comparison on \TimingBatchMatchedBest{} of \TimingN{} datasets (Appendix~\ref{app:timing}).\fi

\subsection{The effect of the backoff distribution}
\label{sec:backoff}
\ifablationready

We compare marginal and continuation backoff with the reference configuration fixed at $K=\KLags$.
Both models are fitted with the same three seeds and training settings.

The three largest backoff effects occur on datasets with low source-specific transition coverage (Section~\ref{sec:coverage}).
Continuation backoff improves MRR by \KnFxReview{} points on \texttt{tgbl-review} and \KnFxComment{} points on \texttt{tgbl-comment} (Figure~\ref{fig:backoff}).
Marginal backoff is stronger on \texttt{Yelp} by \KnLossYelp{} points.
On the remaining datasets, the absolute difference is at most \BackoffRestMax{} points.

Continuation backoff favors destinations reached from many distinct predecessors.
Its gains show the value of this distinction from raw frequency.
Sparse history alone therefore does not determine the better background distribution.

\begin{figure}[t]
\centering
\includegraphics[width=\textwidth]{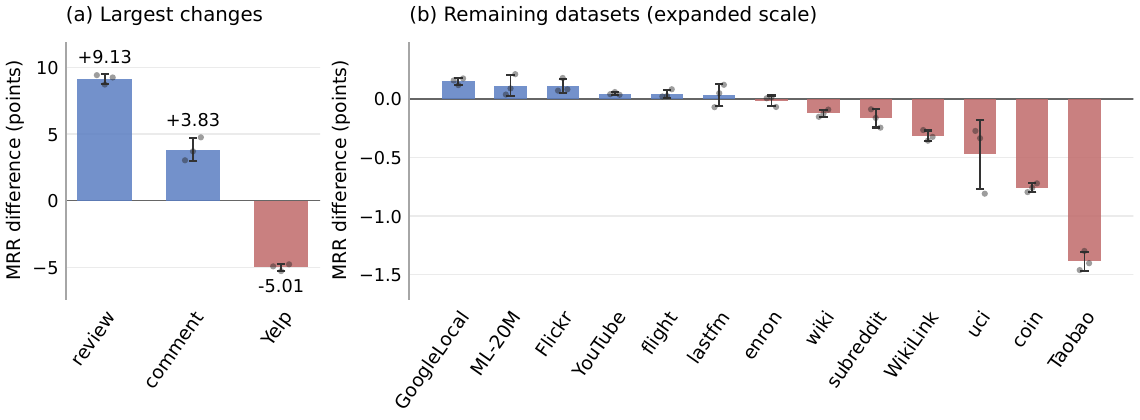}
\caption{Effect of changing marginal to continuation backoff with the reference configuration fixed at $K=\KLags$.
Bars show mean test MRR differences across three paired runs.
Dots show individual runs, and whiskers show one sample standard deviation.
Positive values favor continuation.
Panel (a) shows the three largest absolute mean changes.
Panel (b) uses an expanded linear scale for the remaining datasets.}
\label{fig:backoff}
\end{figure}

\else
\pending{Three-seed backoff comparisons are being recomputed.}
\fi

\subsection{The contribution of co-occurrence and recency}
\label{sec:selection}
\ifablationready

Adding co-occurrence to the reference configuration gives a larger median gain than adding global recency or further recency scales, under either backoff distribution.
With marginal backoff, the median gain is \StageThreeMedianCooc{} MRR points (Appendix~\ref{app:stage3}).
Relative to the reference configuration with the same backoff distribution, the main model gains \SelGainMedian{} points at the median.
Its largest gain is \SelGainMax{} points on \SelGainMaxDataset{}.

The co-occurrence gains suggest that useful associations between destinations extend beyond immediate succession.
Pooling these observations across the recent window supplies evidence that consecutive transition counts omit.

\else
\pending{Three-seed feature comparisons are being recomputed.}
\fi

\subsection{The learned scoring rule}
\label{sec:reading}
\ifablationready

We examine mean coefficients with the reference configuration and marginal backoff on every dataset.
Features differ in scale, so we interpret signs and relative weights within the transition features.
Appendix~\ref{app:weights} reports coefficients for both backoffs and all feature additions.

The source-history coefficient is positive on all \NTGBWeights{} TGB datasets, and recency is positive on \NPosRecTGB{}.
On TGB-Seq, both coefficients are negative on \NNegUserSeq{} of \NSeqWeights{} datasets.
Source history can thus play two roles: its counts can discourage a return to a visited destination, while its recent destinations still provide contexts for predicting a new link through pooled transitions.
The most recent destination has the largest transition coefficient on \NFirstLagLargest{} of \NWeightsTotal{} datasets.

\else
\pending{Three-seed reference coefficients are being recomputed.}
\fi

\subsection{The value of longer contexts}
\label{sec:depth}
\ifdepthready

We fit the reference configuration with marginal backoff separately at each depth from $K=1$ to $K=\KMaxSweep{}$.
Increasing $K$ from one to \KMaxSweep{} improves mean test MRR on all \NSeqDatasets{} TGB-Seq datasets, with gains in every seed (Figure~\ref{fig:ksweep}).
Most of the improvement occurs within the first $\KLags$ destinations.
On these datasets, extending the context from $\KLags$ to \KMaxSweep{} adds at most \DepthSeqBeyondFiveMax{} MRR points.
On TGB, the same increase from one to \KMaxSweep{} improves mean MRR on \NDepthImproveTGB{} of \NTGBDatasets{} datasets.
The smaller gains beyond $K=\KLags$ and the mixed TGB results favor a short recent history as a practical default.
Appendix~\ref{app:ksweep} gives the sweep protocol.

\begin{figure}[t]
\centering
\includegraphics[width=\textwidth]{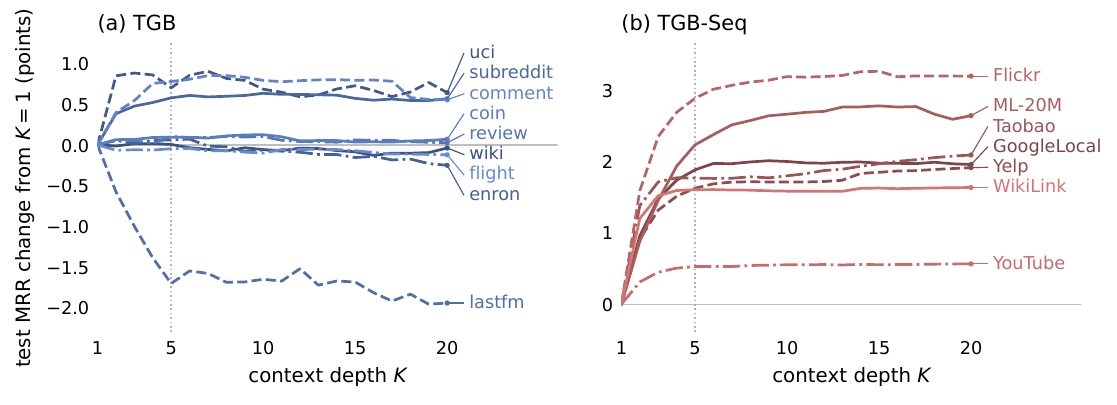}
\caption{Context depth with marginal backoff and the reference configuration.
Lines show mean test MRR changes from $K=1$ across three runs.
The dotted line marks $K=\KLags$.
Panels use different vertical scales.}
\label{fig:ksweep}
\end{figure}

\else
\pending{The three-seed context-depth sweep is being recomputed.}
\fi

\section{When is context available?}
\label{sec:coverage}

Coverage measures whether a context has prior observations when a query is scored.
For a source $u$ with previous destination $v_1$, we distinguish two kinds of evidence.
A pooled transition is available if any source has previously moved from $v_1$ to another destination.
A source-specific transition is available if $u$ has previously done so.
Histories follow the event ordering in Section~\ref{sec:experiments}.

Let $n_u(v_1;t)$ be the decayed count of transitions from $v_1$ contributed by source $u$, so $n(v_1;t)=\sum_{u'} n_{u'}(v_1;t)$.
Here $v_1=v_1(u,t)$ depends on the source and query time.
Over the test queries $Q$, \emph{pooled coverage} is the fraction $\lvert\{(u,t)\in Q:n(v_1;t)>0\}\rvert/\lvert Q\rvert$.
\emph{Pair coverage} is the fraction $\lvert\{(u,t)\in Q:n_u(v_1;t)>0\}\rvert/\lvert Q\rvert$.
The model uses pooled transitions, with exact backoff for empty rows (Equation~\ref{eq:estimators}).

Pooled transition evidence is available on at least \PrevCovMin\% of test queries on every dataset.
By contrast, \NZeroPairSeq{} TGB-Seq datasets have zero pair coverage.
On \texttt{ML-20M}, pooled coverage is complete, so transitions observed for other sources remain available.

\Needspace{20\baselineskip}
\begin{wrapfigure}[18]{r}{\dimexpr\textwidth*11/25\relax}
\centering
\includegraphics[width=\linewidth]{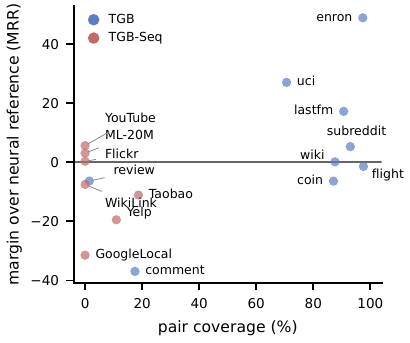}
\caption{Source-specific transition coverage against the MRR margin over the neural reference, one point per dataset.}
\label{fig:coverage}
\end{wrapfigure}

Figure~\ref{fig:coverage} relates context availability to performance.
Among the datasets with zero pair coverage, the model exceeds the reported references on \texttt{ML-20M}, \texttt{Flickr}, and \texttt{YouTube}, but remains below them on \texttt{GoogleLocal} and \texttt{WikiLink}.
At the other end of the coverage range, \texttt{tgbl-coin} and \texttt{tgbl-flight} remain below their references despite pair coverage of at least \PairCovCoin\%.
Context availability alone does not determine performance.

Pair coverage concerns a particular transition context, not the entire source history.
A learned node memory can still encode earlier interactions when pair coverage is zero.
Coverage also differs from the surprise index, which measures whether a test edge occurred during training \citep{poursafaei2022edgebank}.
Suppose a source previously visited $A$ followed by $B$ and has now returned to $A$.
Its next destination $C$ can be new to the source even though the transition context is familiar.
Pair coverage also includes observations after training.

\section{Conclusion}
\label{sec:conclusion}

Our work shows that statistical language modelling can predict temporal links beyond repeated edges.
The proposed model draws on transition and co-occurrence patterns pooled across sources.
These patterns support new links even when the current source has not encountered the same transition before.
In our main evaluation, the model achieves the highest MRR among the compared methods on \NBestAll{} out of \NDatasets{} datasets from TGB and TGB-Seq.
It also outperforms EdgeBank and Base3 on all \NDatasets{} datasets and the heuristic family on \NLearnedAhead{}.
These results show that fixed statistical summaries can provide a competitive alternative to learned node states.

\paragraph{Limitations and future work.}
The timing analysis keeps the configuration and weights fixed (Appendix~\ref{app:timing}).
Future work could test whether choosing marginal or continuation backoff separately for each query improves prediction.

With only \NParams{}--\NParamsFull{} learned parameters, our model provides a simple and competitive baseline against which future neural temporal link predictors should be evaluated.

\label{bodyend}
\clearpage

\subsubsection*{Reproducibility statement}
Tables, figures, and numerical results in the text are generated from the per-run result files and recorded reference scores.
Our count baselines use the same candidate sets as the model.
Code is available at \url{https://doi.org/10.5281/zenodo.22914326}.
We use the public TGB and TGB-Seq distributions with their supplied splits and negative candidates.

\bibliography{refs}
\bibliographystyle{iclr2027/iclr2027_conference}

\appendix

\section{Datasets and reference scores}
\label{app:datasets}

\textbf{Scope.}
We evaluate the \NTGBDatasets{} TGB link-prediction datasets and \NSeqDatasets{} TGB-Seq datasets, each under the shipped candidate sets of its benchmark and evaluator.

\textbf{Sources of the neural reference scores.}
The leaderboards are \url{https://tgb.complexdatalab.com/docs/leader_linkprop/} and \url{https://tgb-seq.github.io/leaderboard/}, last accessed September 18, 2026.
The reference columns contain neural methods.
Counting baselines appear in their separate columns under matched evaluation schedules.
For \texttt{tgbl-wiki} and \texttt{tgbl-review}, we use the leaderboard scores for TPNet and GraphMixer, respectively.
For \texttt{tgbl-coin}, \texttt{tgbl-comment}, and \texttt{tgbl-flight} it is CRAFT, from Tables~2 and 3 of \citet{yi2026future}, evaluated on TGB's shipped negatives with baseline rows that match the leaderboard, and stronger than any leaderboard entry.
For \texttt{tgbl-uci}, \texttt{tgbl-enron}, \texttt{tgbl-lastfm}, and \texttt{tgbl-subreddit} it is TNCN \citep{zhang2025tncn} as reported by \citet{huang2025tgtalker}.

\textbf{Reproduction of EdgeBank.}
We evaluate unlimited memory and the TGB time window, whose duration is fifteen percent of the training time span.
The main tables use event-wise updates.
With batch-200 updates, both variants match the reported scores at their displayed precision on all five current TGB leaderboard datasets.
On \texttt{tgbl-wiki}, unlimited and windowed memory give \TimingEbInfWikiBatch{} and \TimingEbTwWikiBatch{}, compared with the reported \PubEbInfWiki{} and \PubEbTwWiki{}.

\textbf{Reproduction of the heuristic family.}
We implement the four heuristics of \citet{cornell2025heuristics} using the same candidates, tie rule, and event ordering as our model.
This extends their evaluation to \NCornellUnrun{} additional datasets.
Our implementation reproduces their combined score within one point on five of the six TGB datasets they report.
On \texttt{tgbl-review}, their evaluation uses a dataset-specific inverse-recency variant.
The combination stated in their method section obtains \ComboReview{} in our implementation, compared with their reported \CornellReview{}.
The result files include scores for every individual heuristic.
Appendix~\ref{app:timing} compares the same heuristic rule under event-wise and batch-wise updates.

\textbf{Reproduction of Base3.}
We implement Base3 from the released code and evaluate both update schedules.
The main tables use event-wise updates.
With batch-200 updates, our results are within half a point of the reported scores on \texttt{tgbl-comment}, \texttt{tgbl-flight}, and \texttt{tgbl-coin}.
They are lower on \texttt{tgbl-review} and \texttt{tgbl-wiki}.
The released code refers to negative-candidate files that are absent from the repository and specifies paths for a single dataset.
We therefore cannot reconstruct the exact candidates used for those reported results.
Our comparisons use the candidates supplied by TGB.

\textbf{TNCN reference scores.}
For \texttt{tgbl-uci}, \texttt{tgbl-enron}, \texttt{tgbl-lastfm}, and \texttt{tgbl-subreddit}, we use the TNCN results reported by \citet{huang2025tgtalker}.
Their evaluation measures MRR using negatives generated according to the TGB procedure.
Our evaluation uses the candidate files distributed with TGB.
The comparisons therefore share a sampling procedure, but may differ in the particular candidates and the timing of history updates.

\textbf{TGB-Seq references.}
The neural reference scores come from Table~2 of \citet{yi2026future}, which reports CRAFT and SGNN-HN on every TGB-Seq dataset.

\section{The compared methods}
\label{app:baselines}

EdgeBank, the heuristic family, and Base3 are recomputed in our experiments.
The other methods are compared through their reported results.
Appendix~\ref{app:datasets} gives the source of each score.
Our model stores history in count tables with fixed update rules and learns the weights of smoothed estimates and recency features.
The methods below differ in their history representations, update rules, and scoring functions.

\textbf{JODIE} \citep{kumar2019jodie} assigns each user and item a static and a dynamic embedding.
Two recurrent networks update the dynamic embeddings at each interaction using the counterpart, interaction features, and elapsed time.
A learned projection evolves an embedding between interactions, and a linear layer predicts the next item embedding.

\textbf{DyRep} \citep{trivedi2019dyrep} models long-lived associations and transient communications with temporal point processes.
A recurrent architecture updates node embeddings at each event.
A learned intensity over the endpoint embeddings determines the likelihood of an interaction.
Embeddings continue to update during evaluation.

\textbf{TGN} \citep{rossi2020tgn} maintains a memory vector for each node.
Events generate messages that are aggregated and passed to a learned memory updater.
A temporal embedding module combines this memory with neighborhood information.
Memory continues to update during evaluation.

\textbf{SGNN-HN} \citep{pan2020sgnnhn} constructs a graph from the items in a session and connects them through a central node.
A gated GNN propagates information through this graph.
A highway gate combines the initial and updated item embeddings.
An attention-based session representation then scores candidate items.
TGB-Seq retains only nodes seen during training for all compared methods \citep{yi2025tgbseq}.

\textbf{CAWN} \citep{wang2021cawn} samples walks backward in time from both endpoints of a query.
It replaces node identities with occurrence counts relative to the sampled walks.
An RNN encodes the anonymised walks and their time gaps, and an MLP scores the pair from these encodings.

\textbf{EdgeBank} \citep{poursafaei2022edgebank} stores observed source-destination pairs.
It predicts an edge as positive if that pair is in memory.
The unlimited variant retains every observed pair.
The time-window variant retains pairs from a recent interval.
We use the window rule in the TGB reference implementation for this variant.

\textbf{GraphMixer} \citep{cong2023graphmixer} uses an MLP-Mixer to encode recent links with fixed cosine time features.
A separate node encoder averages neighbor features over a time window.
An MLP combines the two representations to score a pair.
The method learns a representation of recent interactions without recurrent networks or self-attention.

\textbf{DyGFormer} \citep{yu2023dygformer} encodes the first-hop interaction sequences of both query nodes.
Features include neighbor attributes, time, and the frequency with which neighbors occur in both sequences.
The sequences are divided into patches and processed by a transformer.
An MLP scores the pair from the resulting representations.

\textbf{PopTrack} \citep{daniluk2023global} maintains decayed destination counts.
It scores a batch from the current counts, then updates and decays them.
The decay factor is selected on validation.

\textbf{TPNet} \citep{lu2024tpnet} maintains random projections of time-decayed temporal walk counts.
Learned MLPs combine the recovered walk features with recent-interaction encodings to score each pair.

\textbf{TNCN} \citep{zhang2025tncn} combines node memory with neural common-neighbor features.
Events update the memories, and a graph transformer computes embeddings from temporal neighborhoods.
The pair score combines endpoint embeddings with information from temporal common neighbors.

\textbf{The heuristic family} of \citet{cornell2025heuristics} uses local and global recency and popularity.
The first statistic in a fixed priority order ranks candidates.
Subsequent statistics break remaining ties.

\textbf{Base3} \citep{kondrup2025base3} combines EdgeBank membership, PopTrack popularity, and a windowed co-occurrence memory.
The co-occurrence component uses decayed neighbor popularity and saturating counts.
Binary confidence indicators determine the interpolation weights through a fixed lookup.

\textbf{CRAFT} \citep{yi2026future} learns node embeddings and uses cross-attention from each candidate to the recent neighbors of the source.
It includes positional information and the time since the candidate was last active.
An MLP produces the final score.
CRAFT-R is its variant for datasets dominated by repeated edges and provides the reference scores on \texttt{tgbl-coin} and \texttt{tgbl-flight}.

\section{The fitted weights}
\label{app:weights}
\ifablationready

Figure~\ref{fig:weights} reports the mean coefficients of the reference configuration under both backoff distributions.
Figure~\ref{fig:weightsopt} reports the weights of co-occurrence, global recency, and further recency scales when each is added separately.
All means use three runs.
A larger weight does not always mean a stronger effect on the score.
Recency and log-probability features have different numerical ranges.
We therefore compare weight sizes only for features of the same kind, such as transition estimates at different history positions.

\begin{figure}[h]
\centering
\includegraphics[width=\textwidth]{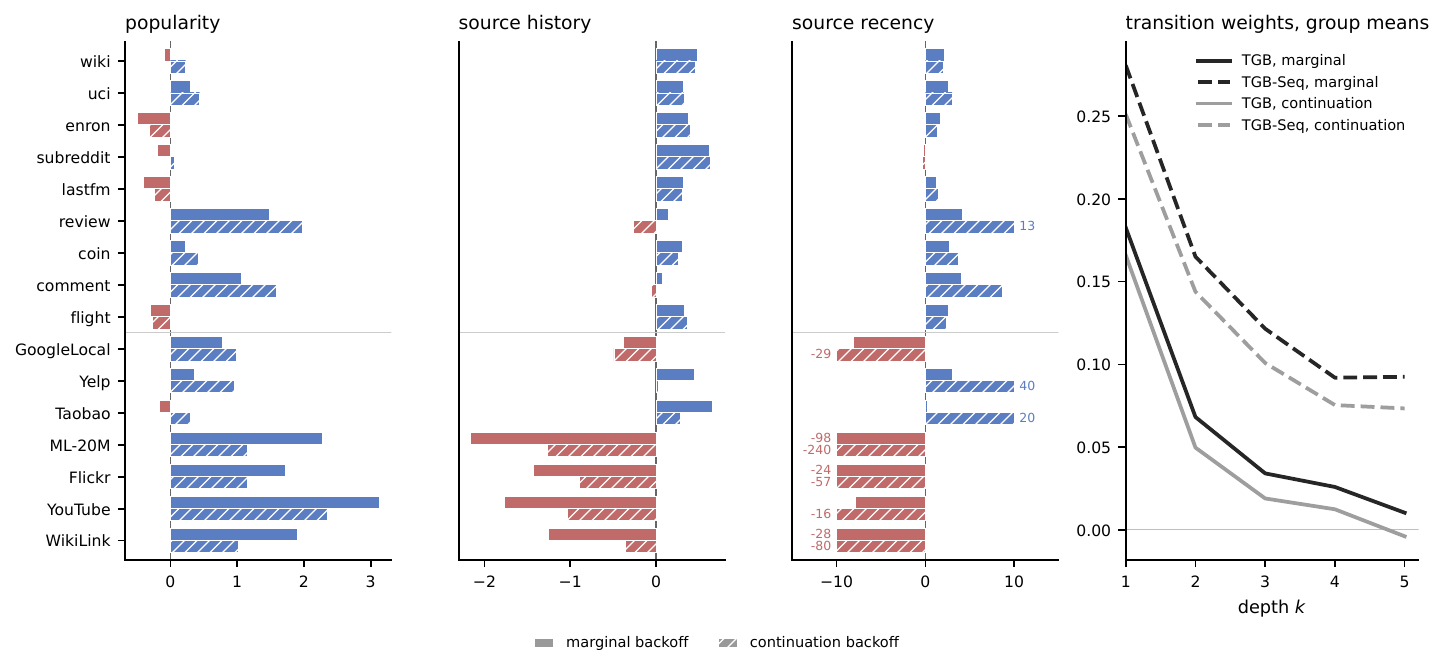}
\caption{Mean fitted weights of the reference configuration over three runs.
Plain bars use marginal backoff and hatched bars use continuation backoff.
TGB datasets appear above the divider.
Blue denotes positive weights and red negative weights.
Recency weights beyond $\pm 10$ are printed beside the clipped bars.
Right: mean transition coefficients at each history position, grouped by benchmark and backoff distribution.}
\label{fig:weights}
\end{figure}

\begin{figure}[h]
\centering
\includegraphics[width=\textwidth]{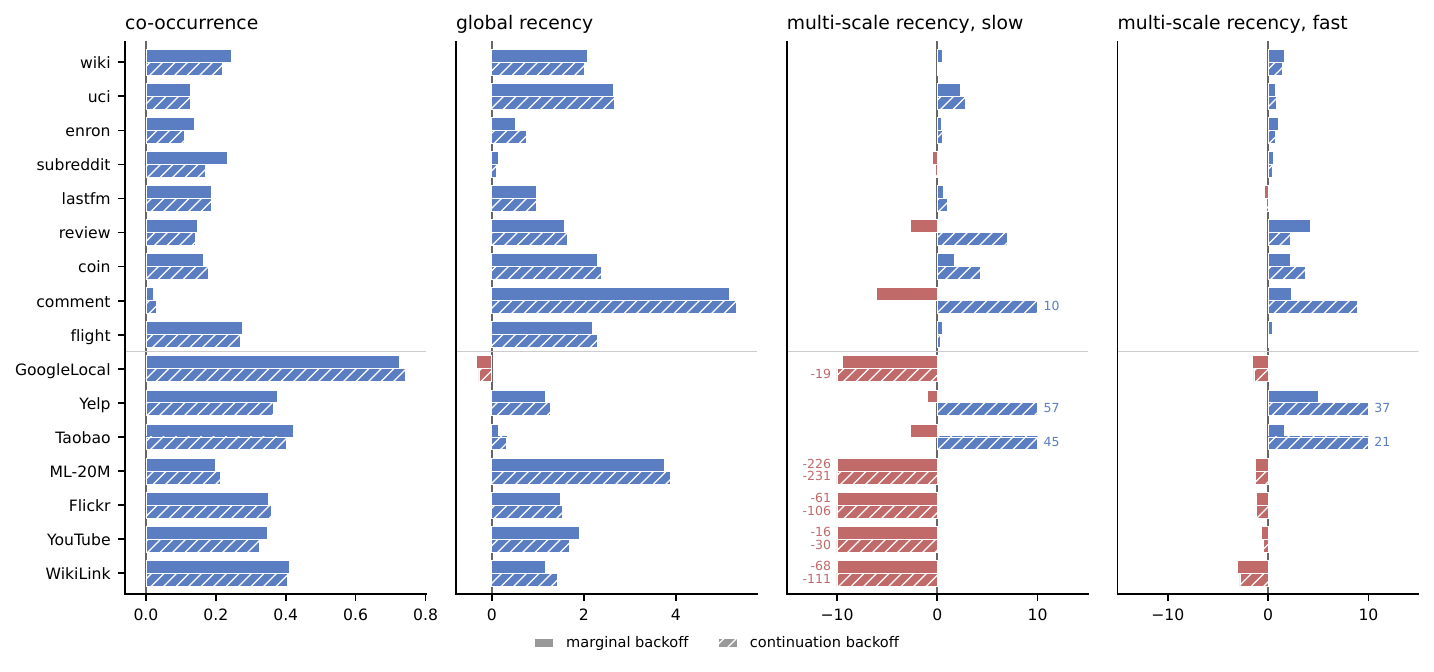}
\caption{Mean weights of co-occurrence and recency features when added to the reference configuration, over three runs.
Backoff distributions follow Figure~\ref{fig:weights}.
The two multi-scale panels show the additional slow ($10\tau_r$) and fast ($\tau_r/10$) recency features.
Weights beyond $\pm 10$ are printed beside the clipped bars.}
\label{fig:weightsopt}
\end{figure}
\FloatBarrier

\else
\pending{Coefficient plots await the three-seed reference and feature fits.}
\fi

\newpage
\section{Sensitivity to the context depth}
\label{app:ksweep}
\ifdepthready

The sweep uses the reference configuration with marginal backoff and three runs on every dataset.
It varies $K$ from one to \KMaxSweep{} while holding the smoothing, decay, and recency settings fixed.
The context depth is fixed at $K=\KLags$ in the main experiments.
A model at depth $k$ uses popularity, source history, source recency, and the first $k$ transition features.
We extract features once at depth \KMaxSweep{}, fit a separate scoring rule for each prefix, and evaluate all depths in one pass.
At $K=\KLags$, we reuse the corresponding reference fit.
Each fit uses every training query, at most 100 epochs, and validation patience 10.
The sweep measures sensitivity within this configuration and does not change the configurations reported in the main tables.
The largest depth equals the smallest sampled-neighbor budget in the TGB-Seq baseline grids, which range from 20 to 60 neighbors \citep{yi2025tgbseq}.

The TGB-Seq gains from $K=1$ to $K=\KMaxSweep{}$ range from \DepthSeqGainMin{} to \DepthSeqGainMax{} MRR points.
The largest gain occurs on \texttt{Flickr}, where source-specific transition coverage is zero (Section~\ref{sec:coverage}).
The model can still draw on earlier destinations because its transition estimates pool observations across sources.
On TGB, mean MRR increases on \texttt{tgbl-uci}, \texttt{tgbl-subreddit}, \texttt{tgbl-review}, \texttt{tgbl-coin}, and \texttt{tgbl-comment}, while it decreases on the remaining datasets.
The largest decline is \DepthLastfmLoss{} points on \texttt{tgbl-lastfm}.

\else
\pending{The marginal reference sweep uses K=1--20 and three seeds.}
\fi

\section{A worked example of the estimators}
\label{app:optionals}

The six-event stream in Figure~\ref{fig:hero} gives a concrete example of each estimator.
We use $\alpha=1$, $K=2$, and undecayed counts.
The vocabulary is $\{a,b,c,d\}$, so $C(t)=6$ and $\lvert\mathcal{V}\rvert=4$.
Candidate $b$ occurs twice, giving $\hat{P}(b)=(2+1)/(6+4)=\ToyPopB$.
Both events belong to $u$, which has five events in total.
The source conditional is therefore $\hat{P}(b\mid u)=(2+\ToyPopB)/(5+1)=\ToyUserB$.

The most recent destination is $v_1=c$.
There is one observed transition from $c$, and it leads to $b$.
Thus $\hat{P}(b\mid c)=(1+\ToyPopB)/(1+1)=\ToyVOneB$.
For an unobserved successor such as $a$, smoothing gives $\hat{P}(a\mid c)=(0+\ToyPopA)/(1+1)=\ToyVOneA$.
The transition row of $d$ is empty, so $\hat{P}(z\mid d)=\hat{P}(z)$ for every candidate.

The continuation base changes this fallback distribution.
Candidate $b$ has followed \ToyContCountB{} distinct predecessors, whereas $c$ has followed \ToyContCountC{}.
Neither $a$ nor $d$ has an observed predecessor.
The total continuation count is \ToyContTotal{}, and $\hat{P}_{\mathrm{c}}(b)=(2+1)/(3+4)=\ToyContB$.
Although $b$ and $c$ occur equally often, the continuation base favors $b$ because it follows more distinct predecessors.
Under continuation backoff, the empty row of $d$ returns $\hat{P}_{\mathrm{c}}$.

Table~\ref{tab:worked} gives all candidate estimates.
The two recent destinations favor different candidates.
The learned score combines these preferences.
Smoothing also assigns positive probability to $d$, which $u$ has never visited.

\begin{table}[!htbp]
\centering
\caption{Smoothed estimates on the worked stream of Figure~\ref{fig:hero}, computed by the model implementation ($\alpha = 1$, counts undecayed).
Conditional rows use marginal backoff.
Bold marks the candidates each estimator favours.}
\label{tab:worked}
\small
\input{tab_worked}
\end{table}

\begin{figure}[t]
\centering
\includegraphics[width=\textwidth]{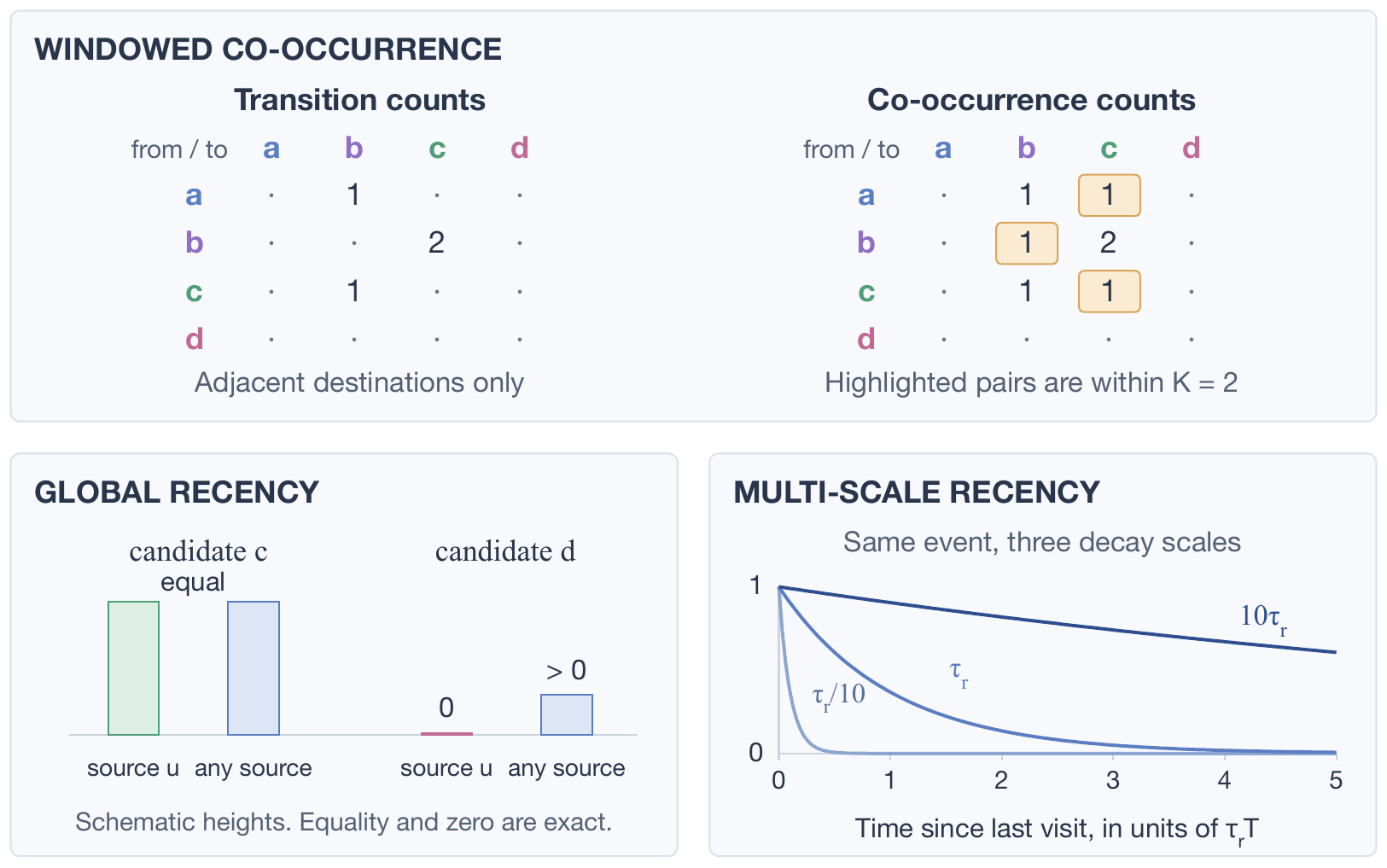}
\caption{Windowed co-occurrence and recency features on the worked stream of Figure~\ref{fig:hero}.
Top: the transition and co-occurrence tables the six events produce.
Each event with prior source history updates one transition cell and up to $K$ co-occurrence cells, one for each previous destination in the window.
Dots mark unobserved pairs.
The highlighted cells, $(a, c)$ from $t_1$ and $t_3$, $(b, b)$ from $t_2$ and $t_4$, and $(c, c)$ from $t_3$ and $t_5$, contain pairs that occur within the window but never consecutively.
Bottom left: the two recency features for candidates $c$ and $d$.
The equality for $c$ is exact because its most recent interaction is with $u$.
The source recency of $d$ is zero because $u$ has never visited it.
Other bar heights are schematic.
Bottom right: the three fixed retention curves of multi-scale recency, independent of the stream.
Their learned weights determine how the time since the source last visited a candidate affects its score.}
\label{fig:optionals}
\end{figure}

Figure~\ref{fig:optionals} illustrates windowed co-occurrence (Equation~\ref{eq:cooccurrence}) and the recency features.
The co-occurrence table records $(a,c)$ from nonconsecutive interactions at $t_1$ and $t_3$.
Global recency distinguishes candidate $d$, which another source visited but $u$ did not.
Multi-scale recency applies three fixed decay scales to the time since the source last visited each candidate.
The features describe complementary aspects of the stream: associations within a window, activity across sources, and time since the source last visited a candidate.

\FloatBarrier

\section{Controlled feature comparisons}
\label{app:stage3}
\ifablationready

We add co-occurrence, global recency, and further source-recency scales to the reference configuration, first separately and then together.
We repeat these comparisons under both backoff distributions, using three runs per configuration.
Figure~\ref{fig:stage3} compares each addition with the reference on the same backoff distribution.
Co-occurrence gives the largest median gain among the individual additions.
With marginal backoff, it improves MRR on \NStageThreeCoocHelps{} of \NMenuTotal{} datasets.
Combining all features has a larger median gain, but improves fewer datasets.
Thus, adding more features does not consistently improve performance.

\begin{figure}[h]
\centering
\includegraphics[width=\textwidth]{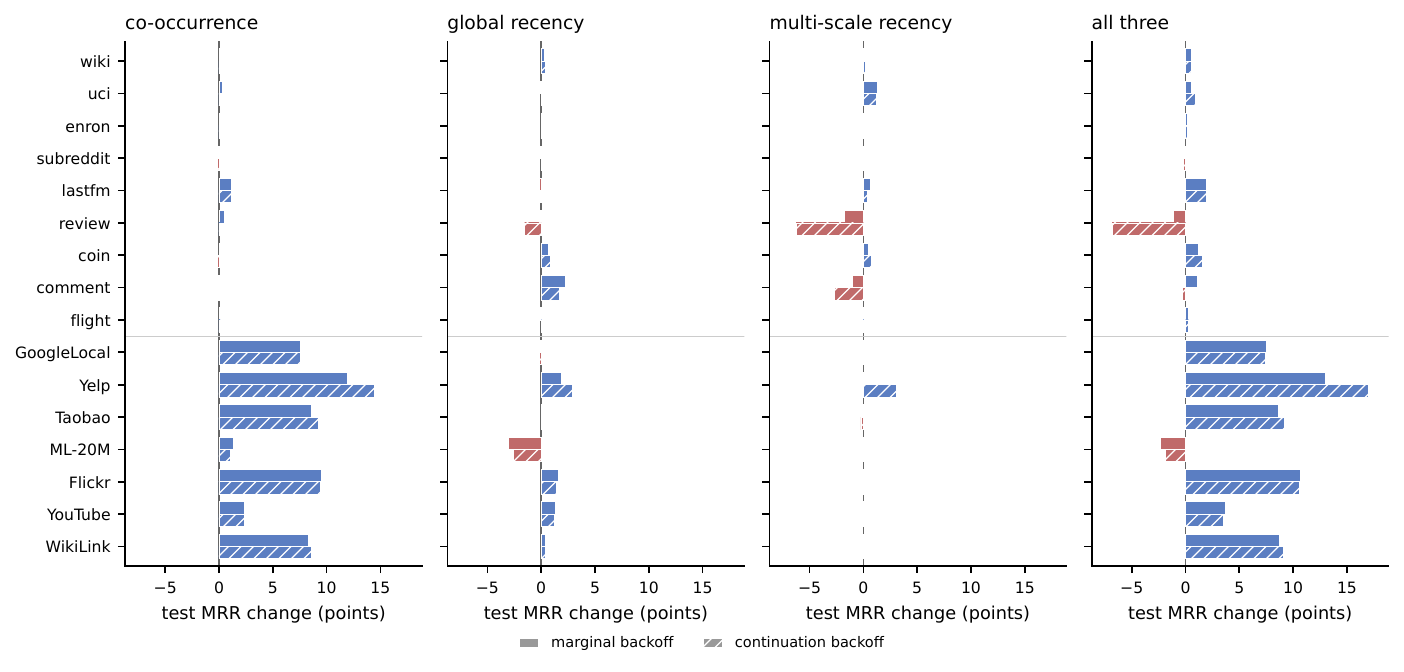}
\caption{Change in mean test MRR from adding co-occurrence, global recency, further source-recency scales, or all three to the reference configuration.
Each addition is compared under the same backoff distribution.
Each comparison uses three runs.
Plain bars use marginal backoff and hatched bars use continuation backoff.
TGB datasets appear above the divider.}
\label{fig:stage3}
\end{figure}
\FloatBarrier

\else
\pending{All ten feature configurations are being evaluated with three seeds.}
\fi

\section{Hyperparameters and their provenance}
\label{app:hyper}
The EdgeBank paper sets the recent-history window to the validation duration \citep[Section~4]{poursafaei2022edgebank}.
We use the same duration for the count-decay constant $\tau$.
Recency uses the training duration $T$.
Each duration is the last timestamp minus the first timestamp in that split.
Both constants are fixed during fitting and evaluation.
The main comparisons use $\alpha=1$.
Appendix~\ref{app:alpha} examines sensitivity to this choice.
The context depth $K=\KLags$ and recency constant $\tau_r=\TauRec$ are also fixed.
Appendix~\ref{app:ksweep} examines context depth, and Appendix~\ref{app:stage3} evaluates additional recency scales.

\begin{table}[htbp]
\centering
\caption{Feature sets used in the main comparisons.
Entries name additions to the reference configuration.
These choices remain fixed across five runs.}
\label{tab:configurations}
\small
\input{tab_configurations}
\end{table}

The reference configuration has $K+4$ learned parameters, including the bias.
Co-occurrence and global recency each add one parameter.
Multi-scale recency adds two, and the configuration with all features has $K+8$.

The two decay constants apply to different quantities.
The count tables use $\tau$, measured in the timestamp units of the data.
The recency feature uses $\tau_r T$, a fixed fraction of the training duration.

The scoring rule is fitted with Adam at learning rate $10^{-2}$ for at most 100 epochs, using batches of 256 queries and one uniform negative per query.
Gradients are clipped to a norm of one.
We use no weight decay.
We measure validation MRR after each epoch and stop after ten consecutive epochs without improvement.
Test evaluation uses the weights from the epoch with the highest validation MRR.
Validation candidates remain fixed across epochs.
Every training event supplies a training query.
Its features are computed before the event updates the count tables.
Every main-model fit stops by the patience criterion before reaching the epoch limit.
\FloatBarrier

\section{Sensitivity to smoothing}
\label{app:alpha}
\ifalphaready

\input{numbers_alpha}
We vary $\alpha$ over $\{10^{-1}, 1, 10\}$ with marginal backoff, the reference configuration, and $K=5$.
The same value is used in the marginal and conditional estimators.
Each setting uses three runs on all 16 datasets.
The $\alpha=1$ results reuse the marginal reference fits from the feature comparisons.
This comparison examines sensitivity within a fixed configuration.
It does not select $\alpha$ for the main results.
Across the three smoothing constants, the median range of mean test MRR is \AlphaRangeMedian{} points.
The largest range is \AlphaRangeMax{} points on \AlphaMaxDataset{} (Figure~\ref{fig:alpha}).

\begin{figure}[!htbp]
\centering
\includegraphics[width=\textwidth]{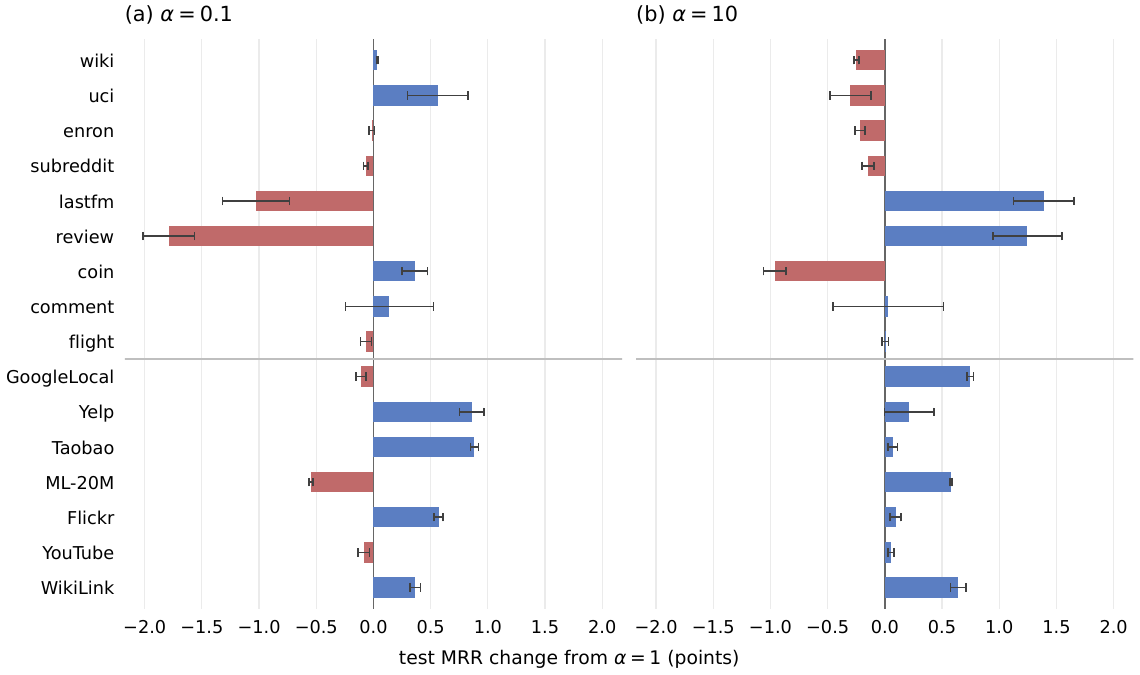}
\caption{Change in test MRR from $\alpha=1$ under smaller and larger smoothing constants.
Bars show mean changes over three seeds.
Error bars show one sample standard deviation of the changes, with runs paired by seed.
Blue denotes increases and red denotes decreases.
TGB datasets appear above the divider.
All settings use marginal backoff, the reference configuration, and $K=5$.}
\label{fig:alpha}
\end{figure}
\FloatBarrier

\newpage

\else
\pending{The three-seed smoothing analysis is being recomputed.}
\fi

\section{Sensitivity to evaluation timing}
\label{app:timing}
\iftimingready

We vary evaluation history while fixing the model configuration and weights.
All conditions use the same five fitted models and test candidates for each dataset.
EdgeBank, Base3, and the heuristic family are also evaluated under event-wise and batch-wise updates.
The heuristic rule remains fixed across schedules.

\emph{Event-wise} evaluation updates history after each scored event, retaining file order for tied timestamps.
\emph{Batch-wise} evaluation scores 200 consecutive test events before adding any of them to history.
\emph{Strict-timestamp} evaluation scores all events at time $t$ using only events with earlier timestamps, including training and validation events.
After scoring a group, we update the tables in file order.
The transition rule is unchanged, but earlier rows at the query timestamp are excluded from its history.

The released EdgeBank and TGB implementations update memory after batches of 200 by default, as in our batch-wise condition.\footnote{ \href{https://github.com/fpour/DGB/blob/main/EdgeBank/link_pred/edge_bank_baseline.py}{EdgeBank code} and \href{https://github.com/shenyangHuang/TGB/blob/main/examples/linkproppred/tgbl-wiki/edgebank.py}{TGB example}.}
TGN also uses earlier test interactions for later predictions, with batching balancing computation and update frequency \citep{rossi2020tgn}.
Computational batches do not always impose this history boundary.
DyGLib models such as TGAT and DyGFormer retrieve neighbors strictly before each query timestamp, including earlier test interactions \citep{yu2023dygformer}.\footnote{ \href{https://github.com/twitter-research/tgn/blob/master/evaluation/evaluation.py}{TGN evaluation}, \href{https://github.com/yule-BUAA/DyGLib_TGB/blob/master/evaluate_link_prediction.py}{DyGLib evaluation}, and \href{https://github.com/yule-BUAA/DyGLib_TGB/blob/master/utils/utils.py}{neighbor sampling}.}
Repeating the event-wise evaluation with the saved weights gives nearly identical test scores.

\begin{table}[!htbp]
\centering
\caption{Test MRR (\%) on TGB under event-wise and batch-200 updates.
EdgeBank, Heuristic, Base3, and Ours share the listed schedule and candidates.
Our scores are means and sample standard deviations over five runs.
Neural reference scores retain their original protocols and are repeated for comparison.
Bold marks the highest MRR in each row.}
\label{tab:timing-tgb}
\small
\input{tab_timing_tgb}
\end{table}

\paragraph{Matched counting baselines.}
Tables~\ref{tab:timing-tgb} and~\ref{tab:timing-tgbseq} report absolute scores under each update schedule.
Under both schedules, our model exceeds EdgeBank and Base3 on all \TimingN{} datasets and the heuristic family on \TimingHeuristicEachAhead{}.
The heuristic is stronger on \texttt{tgbl-uci} and \texttt{tgbl-comment} under event-wise updates, and on \texttt{tgbl-enron} and \texttt{tgbl-comment} under batch-wise updates.
EdgeBank uses the TGB window rule under both schedules (Appendix~\ref{app:datasets}).

\FloatBarrier
\newpage
\begin{table}[!htbp]
\centering
\caption{Test MRR (\%) on TGB-Seq under event-wise and batch-200 updates.
Columns and conventions follow Table~\ref{tab:timing-tgb}.
Neural reference scores retain their original protocols.}
\label{tab:timing-tgbseq}
\small
\input{tab_timing_tgbseq}
\end{table}

With all counting methods evaluated in batches of 200, our model still achieves the highest MRR on \TimingBatchMatchedBest{} of \TimingN{} datasets: \texttt{tgbl-uci}, \texttt{tgbl-subreddit}, \texttt{tgbl-lastfm}, and \texttt{YouTube}, compared with \TimingEventBestAll{} under event-wise updates.

\paragraph{Strict-timestamp sensitivity.}
Table~\ref{tab:timing-timestamp} isolates the effect of excluding earlier events at the query timestamp.
The largest reductions occur on \texttt{tgbl-enron}, \texttt{WikiLink}, \texttt{Flickr}, and \texttt{YouTube}.
Configurations, weights, and candidates remain fixed, so the changes reflect the history available when each query is scored.
Baselines were not evaluated under this schedule.

\begin{table}[!htbp]
\centering
\caption{Strict-timestamp sensitivity of our model.
Scores are test MRR (\%), with means and sample standard deviations over five runs.
Changes from event-wise evaluation use unrounded scores paired by seed.
Configurations, weights, and candidates are fixed.}
\label{tab:timing-timestamp}
\small
\input{tab_timing_timestamp}
\end{table}

\else
\pending{Saved-weight evaluations are pending for event-wise, batch-200, and strict-timestamp updates.}
\fi

\end{document}

%% file: numbers.tex
\newcommand{\pmstd}[1]{{\scriptsize$\pm$#1}}

\newcommand{\BackoffRestMax}{1.39}

\newcommand{\ComboReview}{17.69}

\newcommand{\CornellReview}{52.20}

\newcommand{\DepthLastfmLoss}{1.94}
\newcommand{\DepthSeqBeyondFiveMax}{0.41}
\newcommand{\DepthSeqGainMax}{3.19}
\newcommand{\DepthSeqGainMin}{0.57}

\newcommand{\KLags}{5}
\newcommand{\KMaxSweep}{20}
\newcommand{\KnFxComment}{3.83}
\newcommand{\KnFxReview}{9.13}
\newcommand{\KnLossYelp}{5.01}

\newcommand{\MarginWiki}{0.14}

\newcommand{\NBaseThreeTotal}{16}
\newcommand{\NBeatBaseThree}{16}

\newcommand{\NBehindSeq}{4}
\newcommand{\NBehindTGB}{4}
\newcommand{\NBestAll}{7}

\newcommand{\NBestCraftFamily}{6}

\newcommand{\NCornellUnrun}{10}
\newcommand{\NDatasets}{16}
\newcommand{\NDepthImproveTGB}{5}
\newcommand{\NFirstLagLargest}{16}
\newcommand{\NLearnedAhead}{14}
\newcommand{\NLearnedTotal}{16}
\newcommand{\NMenuTotal}{16}

\newcommand{\NNegUserSeq}{5}

\newcommand{\NParams}{9}
\newcommand{\NParamsFull}{13}
\newcommand{\NPosRecTGB}{8}

\newcommand{\NSeqDatasets}{7}
\newcommand{\NSeqWeights}{7}

\newcommand{\NStageThreeCoocHelps}{14}

\newcommand{\NTGBDatasets}{9}
\newcommand{\NTGBWeights}{9}
\newcommand{\NWeightsTotal}{16}
\newcommand{\NZeroPairSeq}{5}

\newcommand{\PairCovCoin}{87.1}

\newcommand{\PrevCovMin}{96.1}

\newcommand{\PubEbInfWiki}{49.50}

\newcommand{\PubEbTwWiki}{57.10}

\newcommand{\SelGainMax}{13.01}
\newcommand{\SelGainMaxDataset}{\texttt{Yelp}}
\newcommand{\SelGainMedian}{1.44}

\newcommand{\StageThreeMedianCooc}{0.82}

\newcommand{\TauRec}{0.01}

\newcommand{\ToyContB}{0.43}

\newcommand{\ToyContCountB}{2}
\newcommand{\ToyContCountC}{1}

\newcommand{\ToyContTotal}{3}
\newcommand{\ToyPopA}{0.20}
\newcommand{\ToyPopB}{0.30}

\newcommand{\ToyUserB}{0.38}

\newcommand{\ToyVOneA}{0.10}
\newcommand{\ToyVOneB}{0.65}

%% file: result_status.tex
\newif\ifablationready
\ablationreadytrue
\newif\ifdepthready
\depthreadytrue
\newif\ifalphaready
\alphareadytrue
\newif\iftimingready
\timingreadytrue

%% file: numbers_timing.tex
\newcommand{\TimingN}{16}

\newcommand{\TimingEventBestAll}{7}

\newcommand{\TimingBatchMatchedBest}{4}
\newcommand{\TimingHeuristicEachAhead}{14}

\newcommand{\TimingEbInfWikiBatch}{49.47}
\newcommand{\TimingEbTwWikiBatch}{57.10}

%% file: tab_tgb.tex
\begin{tabular}{l rrr rr rl}
\toprule
& & & & \multicolumn{2}{c}{Ours} & \multicolumn{2}{c}{Best neural} \\
\cmidrule(lr){5-6}\cmidrule(lr){7-8}
Dataset & EdgeBank & Heuristic & Base3 & marginal & continuation & MRR & Method \\
\midrule
\texttt{wiki} & 64.40 & 82.06 & 73.26 & \textbf{82.84}\,\pmstd{0.04} & 82.72\,\pmstd{0.10} & 82.70 & TPNet \\
\texttt{uci} & 32.40 & \textbf{52.75} & 35.22 & 51.57\,\pmstd{0.32} & 51.25\,\pmstd{0.26} & 24.50 & TNCN \\
\texttt{enron} & 15.60 & 84.66 & 48.65 & 86.82\,\pmstd{0.01} & \textbf{86.83}\,\pmstd{0.03} & 37.90 & TNCN \\
\texttt{subreddit} & 59.22 & 74.19 & 74.25 & \textbf{74.90}\,\pmstd{0.06} & 74.79\,\pmstd{0.04} & 69.60 & TNCN \\
\texttt{lastfm} & 2.63 & 16.58 & 11.37 & 32.84\,\pmstd{0.18} & \textbf{32.87}\,\pmstd{0.12} & 15.60 & TNCN \\
\texttt{review} & 2.69 & 33.93 & 12.87 & 36.66\,\pmstd{0.74} & 45.76\,\pmstd{0.39} & \textbf{52.10} & GraphMixer \\
\texttt{coin} & 58.31 & 81.10 & 77.71 & 82.09\,\pmstd{0.06} & 81.65\,\pmstd{0.02} & \textbf{88.47} & CRAFT-R \\
\texttt{comment} & 15.03 & 72.38 & 45.41 & 47.94\,\pmstd{1.27} & 54.76\,\pmstd{0.87} & \textbf{91.72} & CRAFT \\
\texttt{flight} & 38.71 & 88.92 & 79.34 & 89.93\,\pmstd{0.06} & 89.99\,\pmstd{0.04} & \textbf{91.39} & CRAFT-R \\
\bottomrule
\end{tabular}

%% file: tab_tgbseq.tex
\begin{tabular}{l rrr rr rl}
\toprule
& & & & \multicolumn{2}{c}{Ours} & \multicolumn{2}{c}{Best neural} \\
\cmidrule(lr){5-6}\cmidrule(lr){7-8}
Dataset & EdgeBank & Heuristic & Base3 & marginal & continuation & MRR & Method \\
\midrule
\texttt{GoogleLocal} & 1.96 & 18.39 & 3.97 & 31.27\,\pmstd{0.02} & 31.42\,\pmstd{0.03} & \textbf{62.88} & SGNN-HN \\
\texttt{Yelp} & 9.77 & 30.81 & 11.99 & 53.21\,\pmstd{0.06} & 52.16\,\pmstd{0.09} & \textbf{72.69} & CRAFT \\
\texttt{Taobao} & 20.28 & 42.99 & 20.17 & 59.58\,\pmstd{0.03} & 58.90\,\pmstd{0.06} & \textbf{70.68} & CRAFT \\
\texttt{ML-20M} & 1.94 & 17.71 & 6.64 & \textbf{39.24}\,\pmstd{0.15} & 39.00\,\pmstd{0.10} & 35.91 & CRAFT \\
\texttt{Flickr} & 1.96 & 38.38 & 34.81 & \textbf{62.68}\,\pmstd{0.07} & 62.66\,\pmstd{0.08} & 62.34 & CRAFT \\
\texttt{YouTube} & 1.96 & 56.64 & 38.09 & \textbf{65.27}\,\pmstd{0.06} & 65.12\,\pmstd{0.04} & 59.64 & SGNN-HN \\
\texttt{WikiLink} & 1.96 & 47.70 & 26.61 & 67.95\,\pmstd{0.05} & 67.97\,\pmstd{0.04} & \textbf{75.48} & CRAFT \\
\bottomrule
\end{tabular}

%% file: tab_worked.tex
\begin{tabular}{lrrrr}
\toprule
Estimator & $a$ & $b$ & $c$ & $d$ \\
\midrule
$\hat{P}(z)$ & 0.20 & \textbf{0.30} & \textbf{0.30} & 0.20 \\
$\hat{P}_{\mathrm{c}}(z)$ & 0.14 & \textbf{0.43} & 0.29 & 0.14 \\
$\hat{P}(z \mid u)$ & 0.20 & \textbf{0.38} & \textbf{0.38} & 0.03 \\
$\hat{P}(z \mid v_1{=}c)$ & 0.10 & \textbf{0.65} & 0.15 & 0.10 \\
$\hat{P}(z \mid v_2{=}b)$ & 0.07 & 0.10 & \textbf{0.77} & 0.07 \\
\bottomrule
\end{tabular}

%% file: tab_configurations.tex
\begin{tabular}{lll}
\toprule
Dataset & Marginal backoff & Continuation backoff \\
\midrule
\texttt{wiki} & all three & all three \\
\texttt{uci} & all three & all three \\
\texttt{enron} & all three & all three \\
\texttt{subreddit} & reference & global recency \\
\texttt{lastfm} & all three & all three \\
\texttt{review} & co-occurrence & co-occurrence \\
\texttt{coin} & all three & all three \\
\texttt{comment} & multi-scale recency & global recency \\
\texttt{flight} & all three & all three \\
\texttt{GoogleLocal} & co-occurrence & co-occurrence \\
\texttt{Yelp} & all three & all three \\
\texttt{Taobao} & co-occurrence & co-occurrence \\
\texttt{ML-20M} & co-occurrence & co-occurrence \\
\texttt{Flickr} & all three & all three \\
\texttt{YouTube} & all three & all three \\
\texttt{WikiLink} & all three & all three \\
\bottomrule
\end{tabular}

%% file: numbers_alpha.tex
\newcommand{\AlphaRangeMedian}{0.75}
\newcommand{\AlphaRangeMax}{3.03}
\newcommand{\AlphaMaxDataset}{\texttt{tgbl-review}}

%% file: tab_timing_tgb.tex
\begin{tabular}{ll rrr r rl}
\toprule
& & & & & & \multicolumn{2}{c}{Best neural} \\
\cmidrule(lr){7-8}
Dataset & Updates & EdgeBank & Heuristic & Base3 & Ours & MRR & Method \\
\midrule
\texttt{wiki} & Event & 64.40 & 82.06 & 73.26 & \textbf{82.84}\,\pmstd{0.04} & 82.70 & TPNet \\
 & Batch 200 & 57.10 & 72.90 & 68.07 & 74.08\,\pmstd{0.09} & \textbf{82.70} & TPNet \\
\addlinespace[3pt]
\texttt{uci} & Event & 32.40 & \textbf{52.75} & 35.22 & 51.57\,\pmstd{0.32} & 24.50 & TNCN \\
 & Batch 200 & 22.16 & 37.49 & 28.22 & \textbf{38.55}\,\pmstd{0.16} & 24.50 & TNCN \\
\addlinespace[3pt]
\texttt{enron} & Event & 15.60 & 84.66 & 48.65 & \textbf{86.83}\,\pmstd{0.03} & 37.90 & TNCN \\
 & Batch 200 & 14.14 & \textbf{54.93} & 45.17 & 52.46\,\pmstd{0.17} & 37.90 & TNCN \\
\addlinespace[3pt]
\texttt{subreddit} & Event & 59.22 & 74.19 & 74.25 & \textbf{74.90}\,\pmstd{0.06} & 69.60 & TNCN \\
 & Batch 200 & 58.85 & 74.18 & 74.14 & \textbf{74.30}\,\pmstd{0.04} & 69.60 & TNCN \\
\addlinespace[3pt]
\texttt{lastfm} & Event & 2.63 & 16.58 & 11.37 & \textbf{32.84}\,\pmstd{0.18} & 15.60 & TNCN \\
 & Batch 200 & 2.57 & 15.74 & 11.32 & \textbf{16.60}\,\pmstd{0.12} & 15.60 & TNCN \\
\addlinespace[3pt]
\texttt{review} & Event & 2.69 & 33.93 & 12.87 & 45.76\,\pmstd{0.39} & \textbf{52.10} & GraphMixer \\
 & Batch 200 & 2.53 & 31.41 & 8.17 & 45.57\,\pmstd{0.39} & \textbf{52.10} & GraphMixer \\
\addlinespace[3pt]
\texttt{coin} & Event & 58.31 & 81.10 & 77.71 & 82.09\,\pmstd{0.06} & \textbf{88.47} & CRAFT-R \\
 & Batch 200 & 57.96 & 80.64 & 76.87 & 81.36\,\pmstd{0.04} & \textbf{88.47} & CRAFT-R \\
\addlinespace[3pt]
\texttt{comment} & Event & 15.03 & 72.38 & 45.41 & 54.76\,\pmstd{0.87} & \textbf{91.72} & CRAFT \\
 & Batch 200 & 14.94 & 72.07 & 44.93 & 54.58\,\pmstd{0.86} & \textbf{91.72} & CRAFT \\
\addlinespace[3pt]
\texttt{flight} & Event & 38.71 & 88.92 & 79.34 & 89.99\,\pmstd{0.04} & \textbf{91.39} & CRAFT-R \\
 & Batch 200 & 38.71 & 88.90 & 79.24 & 89.95\,\pmstd{0.04} & \textbf{91.39} & CRAFT-R \\
\bottomrule
\end{tabular}

%% file: tab_timing_tgbseq.tex
\begin{tabular}{ll rrr r rl}
\toprule
& & & & & & \multicolumn{2}{c}{Best neural} \\
\cmidrule(lr){7-8}
Dataset & Updates & EdgeBank & Heuristic & Base3 & Ours & MRR & Method \\
\midrule
\texttt{GoogleLocal} & Event & 1.96 & 18.39 & 3.97 & 31.42\,\pmstd{0.03} & \textbf{62.88} & SGNN-HN \\
 & Batch 200 & 1.96 & 18.39 & 3.73 & 29.00\,\pmstd{0.01} & \textbf{62.88} & SGNN-HN \\
\addlinespace[3pt]
\texttt{Yelp} & Event & 9.77 & 30.81 & 11.99 & 53.21\,\pmstd{0.06} & \textbf{72.69} & CRAFT \\
 & Batch 200 & 9.76 & 30.81 & 11.88 & 52.26\,\pmstd{0.05} & \textbf{72.69} & CRAFT \\
\addlinespace[3pt]
\texttt{Taobao} & Event & 20.28 & 42.99 & 20.17 & 59.58\,\pmstd{0.03} & \textbf{70.68} & CRAFT \\
 & Batch 200 & 20.27 & 42.99 & 20.16 & 59.57\,\pmstd{0.03} & \textbf{70.68} & CRAFT \\
\addlinespace[3pt]
\texttt{ML-20M} & Event & 1.94 & 17.71 & 6.64 & \textbf{39.00}\,\pmstd{0.10} & 35.91 & CRAFT \\
 & Batch 200 & 1.94 & 17.71 & 6.70 & 31.63\,\pmstd{0.09} & \textbf{35.91} & CRAFT \\
\addlinespace[3pt]
\texttt{Flickr} & Event & 1.96 & 38.38 & 34.81 & \textbf{62.68}\,\pmstd{0.07} & 62.34 & CRAFT \\
 & Batch 200 & 1.96 & 38.35 & 34.54 & 56.61\,\pmstd{0.05} & \textbf{62.34} & CRAFT \\
\addlinespace[3pt]
\texttt{YouTube} & Event & 1.96 & 56.64 & 38.09 & \textbf{65.27}\,\pmstd{0.05} & 59.64 & SGNN-HN \\
 & Batch 200 & 1.96 & 56.48 & 37.61 & \textbf{63.89}\,\pmstd{0.03} & 59.64 & SGNN-HN \\
\addlinespace[3pt]
\texttt{WikiLink} & Event & 1.96 & 47.70 & 26.61 & 67.97\,\pmstd{0.04} & \textbf{75.48} & CRAFT \\
 & Batch 200 & 1.96 & 47.67 & 24.93 & 58.01\,\pmstd{0.02} & \textbf{75.48} & CRAFT \\
\bottomrule
\end{tabular}

%% file: tab_timing_timestamp.tex
\begin{tabular}{l rrr}
\toprule
Dataset & Event-wise & Strict timestamp & Change (points) \\
\midrule
\multicolumn{4}{l}{\textit{TGB}} \\
\texttt{wiki} & 82.84\,\pmstd{0.04} & 82.84\,\pmstd{0.04} & 0.00\,\pmstd{0.00} \\
\texttt{uci} & 51.57\,\pmstd{0.32} & 51.43\,\pmstd{0.32} & -0.14\,\pmstd{0.00} \\
\texttt{enron} & 86.83\,\pmstd{0.03} & 53.50\,\pmstd{0.18} & -33.33\,\pmstd{0.16} \\
\texttt{subreddit} & 74.90\,\pmstd{0.06} & 74.89\,\pmstd{0.06} & 0.00\,\pmstd{0.00} \\
\texttt{lastfm} & 32.84\,\pmstd{0.18} & 32.78\,\pmstd{0.18} & -0.06\,\pmstd{0.00} \\
\texttt{review} & 45.76\,\pmstd{0.39} & 44.93\,\pmstd{0.41} & -0.83\,\pmstd{0.06} \\
\texttt{coin} & 82.09\,\pmstd{0.06} & 81.67\,\pmstd{0.04} & -0.42\,\pmstd{0.03} \\
\texttt{comment} & 54.76\,\pmstd{0.87} & 54.76\,\pmstd{0.87} & 0.00\,\pmstd{0.00} \\
\texttt{flight} & 89.99\,\pmstd{0.04} & 88.94\,\pmstd{0.12} & -1.05\,\pmstd{0.10} \\
\midrule
\multicolumn{4}{l}{\textit{TGB-Seq}} \\
\texttt{GoogleLocal} & 31.42\,\pmstd{0.03} & 31.42\,\pmstd{0.03} & 0.00\,\pmstd{0.00} \\
\texttt{Yelp} & 53.21\,\pmstd{0.06} & 53.21\,\pmstd{0.06} & 0.00\,\pmstd{0.00} \\
\texttt{Taobao} & 59.58\,\pmstd{0.03} & 59.58\,\pmstd{0.03} & 0.00\,\pmstd{0.00} \\
\texttt{ML-20M} & 39.00\,\pmstd{0.10} & 39.00\,\pmstd{0.09} & 0.00\,\pmstd{0.00} \\
\texttt{Flickr} & 62.68\,\pmstd{0.07} & 54.97\,\pmstd{0.07} & -7.71\,\pmstd{0.05} \\
\texttt{YouTube} & 65.27\,\pmstd{0.05} & 61.58\,\pmstd{0.03} & -3.69\,\pmstd{0.05} \\
\texttt{WikiLink} & 67.97\,\pmstd{0.04} & 56.25\,\pmstd{0.04} & -11.72\,\pmstd{0.06} \\
\bottomrule
\end{tabular}